\documentclass[journal]{IEEEtran}
\usepackage{cite,graphicx,amsmath,amssymb,algorithm} 	
\usepackage{textcomp}

\begin{document}


\title{Ensemble and Mixed Learning Techniques for Credit Card Fraud Detection}

\author{Daniel H. M. de Souza and Claudio J. Bordin Jr.\thanks{Universidade Federal do ABC, Santo Andr\'{e}, SP 09210-580 Brazil (e-mail: daniel.henrique@aluno.ufabc.edu.br; claudio.bordin@ufabc.edu.br). This work is licensed under a Creative Commons Attribution-Noncommercial-NoDerivatives license (CC BY-NC-ND 4.0).\\
\includegraphics[width=1cm]{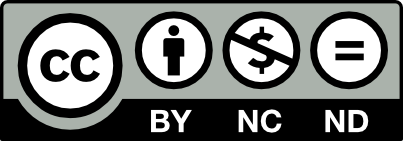}}}

\markboth
{de Souza et al.: Ensemble and Mixed Learning Techniques for Credit Card Fraud Detection}
{de Souza et al.: Ensemble and Mixed Learning Techniques for Credit Card Fraud Detection}


\maketitle
	
\section{Introduction} \label{introducao}

\begin{abstract}
	Spurious credit card transactions are a significant source of financial losses and urge the development of accurate fraud detection algorithms. In this paper, we use machine learning strategies for such an aim. First, we apply a mixed learning technique that uses K-means preprocessing before trained classification to the problem at hand. Next, we introduce an adapted detector ensemble technique that uses OR-logic algorithm aggregation to enhance the detection rate. Then, both strategies are deployed in tandem in numerical simulations using real-world transactions data. We observed from simulation results that the proposed methods diminished computational cost and enhanced performance concerning state-of-the-art techniques.
\end{abstract}

\begin{IEEEkeywords}
	Fraud Detection, Machine Learning, Mixed Learning, Ensembles.
\end{IEEEkeywords}
	
Credit cards are one of the main targets of fraudsters: global losses due to credit card fraud reached \$32 billion in 2020 and are predicted to grow to \$40 billion in 2027~\cite{statistics_fraud_2020}. The most practical means of mitigating such losses is the development of accurate fraud detection models. Current fraud detection algorithms allow credit transactions to be approved or not in real-time according to their match to fraudulent behavior patterns. Supervised machine learning algorithms set up as binary classifiers predominate in this task. Most current models, however, are fragile~\cite{saxena2021notperfect}, as their performances strongly depend on underlying assumptions and hyperparameters.

Previous works on fraud detection deployed diverse machine learning algorithms:~\cite{maes2002fraud} reported a comparison of the performance of Bayesian Networks (BN) and Neural Networks (NN) for fraud detection. Reference~\cite{fraud_tree_nb_2020} considered models based on Decision Trees (DT) and BN. Reference~\cite{shen2007fraud}, in turn, compared NN, DT, and Logistic Regression (LR) models. More recently, to the same aim, \cite{fu2016neuralnet} analyzes the viability of the use of NN, \cite{awoyemi2017ml}  employs Na\"{i}ve Bayes (NB), LR, and K-nearest Neighbors (KNN) models, and \cite{tingfei2020using} introduces a method based on Variational Autoencoding.

Although less commonly, fraud detection setups also deploy unsupervised learning models. That is especially true for configurations with small-scale datasets or unlabeled fraud events. Reference~\cite{lepoivre2016credit} uses K-means (KM) clustering to detect anomalies in credit card transactions. In~\cite{dharwa2011dbscan}, a density-based spatial clustering model (DBSCAN) was used to the same aim. Reference~\cite{sabau_hiearchical_fraud}, in turn, employed a Hierarchical Clustering algorithm, illustrating its usefulness in applications with small datasets. However, for setups with large datasets and labeled fraud events unsupervised models generally lead to inferior performance compared to supervised counterparts.

In this paper, our contribution is twofold. First, we newly deploy a \emph{mixed learning}~\cite{lin_classification_clustering_2017,feng2020evaluation,RAHAMATHUNNISA2020svm_kmeans} technique that performs KM clustering as a preprocessing step before supervised classification to the problem at hand; specifically, we train supervised models separately for each data cluster determined by KM. Second, we introduce a new \emph{ensemble} technique~\cite{ensemble_cancer,ensemble_forecast,ensemble_invasao,coletta2015ensembles,randhawa2018credit} that aggregates the decisions of multiple models using the OR logic function. By combining both methods in tandem, we developed novel detection models that improved performance concerning state-of-the-art models using actual credit card transaction data. We performed simulations in which we implemented and evaluated the fraud detection performance of the machine learning models KNN, NB, LR, Random Forest (RF), Gradient-Boosted Tree (GBT), and Multilayer Perceptron (MLP), individually, in ensembles, and in mixed learning configurations. As far as we know, previous fraud detection setups did not consider the simultaneous use of ensembles and mixed learning.

The remainder of this paper is structured as follows. Section~\ref{referencial_teorico} briefly reviews the machine learning algorithms used in this work, discusses the class imbalance problem, and describes ensemble and mixed learning techniques. Next, Section~\ref{algoritmos_propostos} introduces the newly proposed methods. Section~\ref{simulacoes}, in turn, reports the setup and results of numerical simulations that evaluate the performance of the proposed methods. Finally, we leave our conclusions to Section~\ref{conclusao}.
	
\section{Current Machine Learning Methods} \label{referencial_teorico}

In this section, we briefly review the machine learning algorithms used in this work, discuss the issue of class imbalance and describe ensemble and mixed learning techniques.

\subsection{Machine Learning Algorithms} \label{algoritmos_individuais}

The KNN algorithm~\cite{awoyemi2017ml} is nonparametric and aims to classify a new object given a training dataset correctly labeled for classes. To this intent, the algorithm calculates a distance measure between the new object and each of the training objects to find the $K$ nearest ones, dubbed neighbors. The class of the new object is then determined as the majority class of the $K$ neighbors.

The NB algorithm~\cite{fraud_tree_nb_2020} is a Bayesian classifier. It determines posterior probabilities of objects belonging to each class according to the Bayes'\@ law, operating however under the simplifying assumption that the class features are conditionally independent given the class label.

The LR algorithm~\cite{yang2020optimization_hyperparameter} is a multivariate statistical model for approximating the probability of an object belonging to a class given a set of explanatory variables. Its output variable results from the nonlinear logistic transformation and can be interpreted as a ratio of probabilities. Despite the nonlinearity of the LR model, its coefficients can be estimated via maximum likelihood techniques~\cite{yang2020optimization_hyperparameter}.

NN are algorithms that supposedly mimic human cognition which can be used to recognize patterns in a dataset. They are composed of artificial neurons disposed in layers~\cite{fu2016neuralnet}. An MLP network is composed of at least three layers: an input, a hidden, and an output layer. Each network node (barring the input ones) applies a generally nonlinear activation function to its output, which can be affected or not by an adaptable bias. An MLP seeks to determine the weights and biases that minimize the error in the generated outputs, which can be accomplished via backpropagation algorithms~\cite{fu2016neuralnet}. 

The DT algorithm~\cite{yang2020optimization_hyperparameter} can be used for classification or regression. Its principle is to partition data, attributing objects to nodes according to some optimality criterion such that the resulting tree is the smallest possible one. Most training strategies employ a divide and conquer method, partitioning data in subsets until all nodes are \emph{pure} (i.e., contain data of a single class) or until some stop criterion is reached. Proposals of new data partitions are executed according to the brought \emph{information gain}, calculated according to some metric. DTs are the base of other algorithms such as the RF~\cite{Xuan2018rf} and Boosted DT~\cite{yang2020optimization_hyperparameter}.

The RF and GBT algorithms are modifications of the DT algorithm. The RF algorithm builds a set of separate decision trees, each using a subset of training objects obtained by sampling with replacement from the complete training set via bootstrap~\cite{boostrap_sample}. Then, the majority vote among all trees determines the final classification. The GBT model, in turn, sequentially combines DTs so that the residues of previous ones determine subsequent DTs~\cite{yang2020optimization_hyperparameter}.

\subsection{Class Imbalance}\label{unbal}

A general issue regarding fraud detection is that the number of fraudulent transactions is generally far smaller than the number of genuine ones. Therefore, fraud detection is a classification problem in which one of the classes is far less frequent than the other, which leads most existing machine learning models to exhibit poor performance~\cite{awoyemi2017ml}. One possible solution is to \emph{rebalance}~\cite{awoyemi2017ml,goyal2021supersampling} the training set by sampling with replacement from the minority class, so that, in the new training set, classes are balanced. Another solution is to modify machine learning models so that they can properly deal with imbalanced classes.

For LR and RF models, \cite{LUO2019classweight} proposes a modification to the log-likelihood function, attributing to each class a weight inversely proportional to its relative frequency during the majority vote step. That method can be generalized to other models like GBT~\cite{ALSHARKAWI2021classweightgbt} and MLP~\cite{HUANG2016mlpclassweight}. We propose here to extend it to KNN models.

As we discuss in Section~\ref{mixed_learning_teoria}, mixed learning techniques can also be used to remedy class imbalance.

\subsection{Ensembles} \label{combinando_classificadores}

An \textit{ensemble}~\cite{hruscka_ensembles} is defined as a combination of machine learning algorithms. The composition of an ensemble of similar models is generally not adequate~\cite{coletta2015ensembles}, as it may lead to mass prediction errors. To avoid this effect, one may combine the outputs of different models such as LR, DT, KNN, and NN, among others. The decision of an ensemble is determined via some aggregation method, of which majority vote is the most common~\cite{coletta2015ensembles}.

Ensembles have been used with advantage in several supervised learning problems, such as cancer detection~\cite{ensemble_cancer}, detections of intrusion in computer networks~\cite{ensemble_invasao}, text mining~\cite{hruscka_ensembles}, and time-series forecast~\cite{ensemble_forecast}. Algorithm~\ref{alg1} describes an ensemble using majority vote (CC-MV).

\begin{algorithm}
				1. \textbf{for} $i=1$ \textbf{to} $n$:
				\\
				1.1. \textbf{for} $j=1$ \textbf{to} $m$:\\
				1.1.1. Classify object $i$ using the model $M_{j}$. Define the result as $y_{ij}$\\
				1.2  \textbf{end for}\\
				1.3. $y_{i} = \text{mode}\{y_{i1}, y_{i2}, \cdots, y_{im}\}$\\
				2. \textbf{end for}\\
				3. \textbf{return} $y = \{y_{1}, y_{2}, \cdots, y_{n}\}$
\caption{Ensemble using Majority Vote (CC-MV)\label{alg1}}
\end{algorithm}

\subsection{Mixed Learning} \label{mixed_learning_teoria}

Mixed learning \cite{mixed_learning_oldRef} consists of combing unsupervised and supervised learning techniques. In this work, we employ clustering as a preprocessing step so that the dataset is partitioned in clusters and the objects of each cluster are separately processed by a supervised classifier (Algorithm~\ref{alg2}). As discussed in~\cite{coletta2015ensembles}, this procedure can be justified by the fact that similar objects tend to belong to the same class. Therefore, grouping similar objects can improve the generalization capability of a classifier, which is particularly useful when data of one class is scarce as in problems with imbalanced classes.

Other works deploy mixed learning in diverse problems, especially in setups that demand large training sets and have imbalanced classes. Reference~\cite{lin_classification_clustering_2017} introduces a mixed learning technique for breast cancer detection, which uses KM to partition the training set in $K=5$ clusters and MLP and DT for classification; this method improved classification performance compared to algorithms that do not use clustering. Reference~\cite{feng2020evaluation} uses KM for the clustering of food quality data and, subsequently, employs Support Vector Machines (SVM)~\cite{awad2015support} for food classification. The work in~\cite{RAHAMATHUNNISA2020svm_kmeans} performs clustering via KM and classification using SVM for detecting diseases in plants. Reference~\cite{bablani2020kneams_nn}, in turn, clusters EEG data via KNN and performs classification via NN. In~\cite{almantara2020dbscan_knn}, on the other hand, spatial data are grouped via the DBSCAN algorithm and later classified using KNN.

\begin{algorithm}
			1. Using clustering, partition the dataset in $n$ clusters\\
			2. \textbf{for} $i=1$ \textbf{to} $n$:	\\
			2.1. Classify the objects in the cluster $i$ using a supervised classification model, with result $y_{i}$ 
			\\
			3. \textbf{end for}\\
			4. \textbf{return} $y = \{y_{1}, y_{2}, \cdots, y_{n}\}$\\
		\caption{Mixed Learning\label{alg2}}
\end{algorithm}

\section{Proposed Methods} \label{algoritmos_propostos}

\subsection{Ensembles using the OR logic} \label{novo_metodoensemble_funcaoOR}

As discussed in Section~\ref{combinando_classificadores}, ensembles can mitigate the effects of class imbalance. Most commonly, ensembles determine a decision by the majority vote in a set of models. We propose here an alternative aggregation method, applicable to binary detection problems, referred to as CC-OR (Classifier Combination Using OR Logic), in which detection of fraud by a single member (taken as logic level `1') suffices for the ensemble to classify the event as fraudulent. 

Empirically, we verified that CC-OR increases fraud detection rate without a relevant increase of false positives. An additional advantage of CC-OR is computational complexity savings in relation to previous aggregation methods, as CC-OR dispenses with the need to evaluate the mode~\cite{complexidade_moda} of the decisions of each model, and that, after a positive detection by a single model, the execution of the remaining ones can be halted.

Algorithm~\ref{alg3} describes the proposed CC-OR ensemble method.

\begin{algorithm}
		1. \textbf{for} $i=1$ \textbf{to} $n$:\\
		1.1. \textbf{for} $j=1$ \textbf{to} $m$:\\
		1.1.1 Classify the object $i$ using the classifier $M_{j}$, defining the result as $y_{ij}$ \\
		1.2. \textbf{end for}\\
		1.3. \textbf{if} $\sum_{j=1}^{m} y_{ij}\geq 1$ \textbf{then} $y_{i} = 1$\\
		1.4 \textbf{else} $y_{i}=0$\\
		1.5. \textbf{end if}\\
		2. \textbf{end for}\\
		3. \textbf{return} $y = \{y_{1}, y_{2}, \cdots, y_{n}\}$
		\caption{Ensemble using the OR Logic (CC-OR)\label{alg3}}
\end{algorithm}

\subsection{Classification Algorithms CCK-MV and CCK-OR}  \label{novo_metodo_PRINCIPAL}

As previously discussed, mixed learning and ensemble strategies help mitigate the effects of class imbalance observed in fraud detection problems. We propose to deploy both methods in tandem, i.e., use an ensemble of classifiers in a mixed learning setup, as described in Algorithm~\ref{alg4}.

\begin{algorithm}
				1. Using a clustering model, partition the dataset in $k$ clusters\\
				2. \textbf{for} $c=1$ \textbf{to} $k$:
				\\
				2.1. \textbf{for} $i=1$ \textbf{to} $n$:
				\\			
				2.1.1. \textbf{for} $j=1$ \textbf{to} $m$:\\	
				2.1.1.1. Classify each object $i$ of cluster $c$ using each individual classifier $M_{cj}$, being the result defined as $y_{cij}$\\
				2.1.2. \textbf{end for}\\
				2.1.3. Combine the classifiers outputs, leading to the combined output $y_{ci}$\\
				2.2. \textbf{end for}\\
				2.3. Join the outputs as $y_{c} = \{y_{c1}, y_{c2}, \cdots, y_{cn}\}$ \\
				3. \textbf{end for}\\
				4. \textbf{return} $y = \{y_{1}, y_{2}, \cdots, y_{k}\}$
		\caption{Combination of Mixed Learning and Ensembles\label{alg4}}		
\end{algorithm}

The proposed setup can deploy different clustering and ensemble
aggregation methods.  In this work, we exclusively use the KM clustering model due to its linear computational cost~\cite{feng2020evaluation}, contrasted to the quadratic cost of hierarchical~\cite{sabau_hiearchical_fraud} and DBSCAN~\cite{complexidade_computacional_clustering} clustering algorithms. The DBSCAN algorithm was also not considered for its discarding of outliers, which is problematic for fraud detection. Thus, the methods described in the sequel only differ in the choice of the ensemble aggregation strategy. The first method (CCK-MV), described in Algorithm~\ref{alg5}, uses majority vote and the second method (CCK-OR), shown in Algorithm~\ref{alg6}, uses the CC-OR aggregation strategy.

\begin{algorithm}
				1. Using \textit{K-means}, partition the dataset in $k$ clusters\\
				2. \textbf{for} $c=1$ \textbf{to} $k$:
				\\
				2.1. \textbf{for} $i=1$ \textbf{to} $n$:
				\\			
				2.1.1. \textbf{for} $j=1$ \textbf{to} $m$:\\	
				2.1.1.1. Classify each object $i$ of cluster $c$ using each individual classifier $M_{cj}$, leading to the result $y_{cij}$\\
				2.1.2. \textbf{end for}\\
				2.1.3. $y_{ci} = \text{mode}\{y_{ci1}, y_{ci2}, \cdots, y_{cim}\}$\\ 
				2.2. \textbf{end for}\\
				2.3. Join the outputs as $y_{c} = \{y_{c1}, y_{c2}, \cdots, y_{cn}\}$ \\
				3. \textbf{end for}\\
				4. \textbf{return} $y = \{y_{1}, y_{2}, \cdots, y_{k}\}$
\caption{CCK-MV\label{alg5}}	
\end{algorithm}

\begin{algorithm}
				1. Using \textit{K-means}, partition the dataset in $k$ clusters\\
				2. \textbf{for} $c=1$ \textbf{to} $k$:
				\\
				2.1. \textbf{for} $i=1$ \textbf{to} $n$:
				\\			
				2.1.1. \textbf{for} $j=1$ \textbf{to} $m$:\\	
				2.1.1.1. Classify each object $i$ of cluster $c$ using each individual classifier $M_{cj}$, leading to the result $y_{cij}$\\
				2.1.2. \textbf{end for}\\
				2.1.3. \textbf{if} $\sum_{j=1}^{m} y_{cij}\geq 1$ \textbf{then} $y_{ci} = 1$\\
				2.1.4. \textbf{else} $y_{ci}=0$\\
				2.1.5. \textbf{end if}\\
				2.2. \textbf{end for}\\
				2.3. Join the outputs as $y_{c} = \{y_{c1}, y_{c2}, \cdots, y_{cn}\}$ \\
				3. \textbf{end for}\\
				4. \textbf{return} $y = \{y_{1}, y_{2}, \cdots, y_{k}\}$
				\caption{CCK-OR\label{alg6}}
\end{algorithm}
	
\section{Experiments}
\label{simulacoes}
	
In this section, we present the results of numerical simulations in which we compare the performance of the proposed algorithms with that of previous state-of-the-art ones. In Section~\ref{pre_processamento_implementacao}, we detail the preprocessing, variable selection, and hyperparameter optimization procedures and the implementation of individual models and ensembles. In Section~\ref{dtset}, in turn, we describe the datasets used for performance evaluation. Finally, in Section~\ref{replica_Awoyemi}, we present the simulation results.
	
\subsection{Data Preprocessing and Implementation of Models} \label{pre_processamento_implementacao}
	
\subsubsection{Data Preprocessing} \label{modelagem_pre_processamento}

Initially, the dataset features were normalized by the \emph{min\_max\_scaler} method~\cite{min_max_scaler_spam}, in which the value of the feature of each object is mapped onto the $[0\;1]$ interval. Next, for the mixed learning techniques, the datasets were partitioned into $K=\{2, 3, 4, 5\}$ clusters using KM. Finally, if required, class rebalancing (Section~\ref{unbal}) was performed.
	
For the selection of variables, we used a combination of \emph{filter} and \emph{wrapper}~\cite{feature_selection_general,feature_importance} methods. Given the large datasets used in this work, we used the Pearson Correlation and Mutual Information~\cite{feature_selection_general} filter methods and the Feature Importance~\cite{feature_importance} wrapper method, being the latter evaluated via the results of an RF. For each of those selection methods, the relevance of each variable is inferred and compared to the benchmarks in~\cite{benchmark_pearson_correlation,benchmark_mutual_information,funcionamento_featureimportance} to obtain each decision. Then, each feature is classified as relevant or not by the majority vote among the three models. 
	
\subsubsection{Simulations Setup} \label{implement_classifiers}

In a first experiment, we implemented the individual models KNN, LR, RF, GBT, and MLP using their classical formulations (Section~\ref{algoritmos_individuais}) and their versions adapted to deal with class imbalance (Section~\ref{unbal}). For the MLP, we ran each variant with the optimization functions \emph{Adam}-Stochastic Gradient~\cite{mlp_adam} and \emph{lbfgs}-Quasi-Newton~\cite{quasi_newton_mlp}. The classical models were trained and tested with rebalanced datasets, while the adapted ones received the original datasets. We also implemented the ensemble strategies CC-MV and CC-OR. For the CC-MV, we tested all possible ensembles\footnote{Note that, to avoid ties in majority voting, the number of combined methods for the CC-MV strategy must be odd.} with 3 and 5 individual models, in a total of 1573 distinct ensembles. For the CC-OR, in turn, we tested all possible ensembles with 2, 3, 4, and 5 individual models, in a total of 2366 ensembles. The results of this first experiment are shown in Section~\ref{replica_Awoyemi}.

In a second experiment, we implemented mixed learning techniques. The supervised classification step employed individual models, the CCK-MV (Algorithm~\ref{alg5}), and the CCK-OR (Algorithm~\ref{alg6}) algorithms. The hyperparameters~\cite{yang2020optimization_hyperparameter} were determined via an exhaustive search procedure as follows: a set of hyperparameter values was selected and, next, the methods were trained and tested for each possible value combination, being selected the values that maximized the \emph{F1 score}, defined as
\begin{equation}
F1 = \frac{T_{P}}{T_{P} + \frac{1}{2}\left(F_{P} + F_{N}\right)},
\end{equation}
\noindent using \emph{holdout}~\cite{yadav2016analysis}, where $T_P$, $F_{P}$, and $F_{N}$ a stand for true positive, false positive and false negative rates, respectively. 

For the validation of the hyperparameters, the models were trained and tested via K-Fold~\cite{yadav2016analysis} using K = 10. By this method, each model was trained and tested 10 times; each run employed 9/10 of the training set for training and 1/10 for testing. For performance assessment, we followed~\cite{awoyemi2017ml}, evaluating for each model run the following metrics: Accuracy (acc), Balanced Classification Rate (bcr), Sensitivity (sens), and Specificity(spec), defined as
\begin{eqnarray}
\displaystyle\text{acc} &=& \frac{T_{P} + T_{N}}{T_{P} + T_{N} + F_{P} + F_{N}},\\
\displaystyle\text{sens}&=& \frac{T_{P}}{T_{P} + F_{N}},\\
\displaystyle\text{spec}&=& \frac{T_{N}}{T_{N} + F_{P}},\\
\displaystyle\text{bcr} &=& \frac{1}{2} \cdot \left(\frac{T_{P}}{T_{P} + F_{N}} + \frac{T_{N}}{T_{N} + F_{P}}\right) \nonumber \\
							&=& \frac{1}{2} \cdot \left(\text{sens} + \text{spec} \right),
\end{eqnarray}
\noindent where $T_N$	denotes the true negative rate.

\subsection{Datasets}\label{dtset}

To evaluate the performance of the fraud detection algorithms, we employed two datasets. The first dataset~\cite{mlgroup2017fraud} contains information about actual credit card transactions that occurred in Europe in September of 2013. This dataset is imbalanced, with a total of 284,807 transactions, in which 284,315 (99.83\%) are genuine, and 492 (0.17\%) are fraudulent. It has a total of 31 variables: the transactions timestamps, the class label (1 for fraud and 0 otherwise), and 29 additional explaining variables. This dataset was made available in two parts, being one of them for training/\-testing, with 227,485 transactions (227,455 genuine and 390 fraudulent) and the other, containing 56,962 observations (56,860 genuine e 102 fraudulent), for validation.

The second dataset~\cite{yap2020turca} was provided by the Turkish company Yapi Kredi Teknoloji and contains a set of 62,380 transactions, of which 61,572 (98.70\%) are genuine, and 808 (1.30\%) are fraudulent. This imbalanced dataset has 26 variables: the transactions timestamps, the class label (1 for fraud and 0 otherwise), and 24 additional explaining variables. This dataset was also made available in two parts, being one of them for training/\-testing, with 31,190 transactions (30,400 genuine and 790 fraudulent) and the other, containing 31,190 observations (31,172 genuine e 18 fraudulent) for validation. Despite the transactions represented in this dataset being actual, the variables are numeric, resulting from preprocessing by Principal Component Analysis (PCA)~\cite{MINKA2001PCA}, and do not have a description due to confidentiality concerns.
	
\subsection{Simulation Results}\label{replica_Awoyemi}

\subsubsection{Individual Models vs. Ensembles} \label{results_ensembles}

We ran the first experiment described in Section~\ref{implement_classifiers} and tabulated the resulting performance metrics for individual models and the three best ensembles using both CC-OR and CC-MV. The tables are first sorted by the bcr metric, next by sens, and finally by the mean of all metrics. To facilitate the interpretation of the results, Table~\ref{siglas_modelos} lists all employed acronyms. In the sequel, we refer to methods as \emph{good performing} if their correct classification rates (CCRs) are of at least 70\%~\cite{benchmark_good_model}; our comments focus on those methods.

Tables~\ref{tab_result_ensemble_europeia} and \ref{tab_result_ensemble_turca} list, respectively, the results obtained using the validation
sets of the European and Turkish databases described in Section~\ref{dtset}. The numbers after the CC-MV and CC-OR ensembles are the (arbitrary) indexes of the best-performing configurations described in Table~\ref{tab_lista_ensembles}.

\begin{table} [htb]
	\caption{Acronyms used in the Result Tables.}
	\label{siglas_modelos}
	\begin{center}
		\begin{tabular}{|l|l|}
			\hline
			Model & Acronym \\ \hline
			Na\"{i}ve Bayes & NB \\ \hline
			K-Nearest Neighbors & KNN \\ \hline
			Modified K-Nearest Neighbors & KNN-m \\ \hline
			Logistic Regression & LR \\ \hline
			Modified LR & LR-m \\ \hline
			Random Forest & RF \\ \hline
			Modified Random Forest & RF-m \\ \hline
			Gradient Boosted Tree & GBT \\ \hline
			Modified Gradient Boosted Tree & GBT-m \\ \hline
			MLP with Adam optimizer & MLP-A \\ \hline
			Modified MLP with Adam optimizer & MLP-A-m \\ \hline
			MLP with lbfgs optimizer & MLP-l \\ \hline
			Modified MLP with lbfgs optimizer & MLP-l-m \\ \hline
		\end{tabular}
		
	\end{center}
\end{table}

\begin{table} [htb]
	\caption{Performance of Individual Models and Ensembles for the European Transactions Database~\cite{mlgroup2017fraud}.}
	\label{tab_result_ensemble_europeia}
	\begin{center}
		\begin{tabular}{|l|l|l|l|l|l|}
			\hline
			Model/Ensemble & acc & bcr & sens & spec & Mean \\ \hline
			CC-MV 8 & 0.998 & 0.911 & 0.824 & 0.999 & 0.910 \\ \hline
			CC-MV 9 & 0.998 & 0.911 & 0.824 & 0.999 & 0.910 \\ \hline
			CC-MV 16 & 0.998 & 0.911 & 0.824 & 0.999& 0.910 \\ \hline
			CC-OR 14 & 0.998 & 0.911 & 0.824 & 0.999& 0.910 \\ \hline
			CC-OR 15 & 0.998 & 0.911 & 0.824 & 0.999 & 0.910 \\ \hline
			CC-OR 24 & 0.998 & 0.911 & 0.824 & 0.999 & 0.910 \\ \hline
			KNN & 0.998 & 0.897 & 0.794 & 0.999  & 0.905 \\ \hline
			KNN-m & 0.998 & 0.897 & 0.794 & 0.999 & 0.905 \\ \hline
			LR & 0.998 & 0.892 & 0.784 & 0.999 & 0.898 \\ \hline
			LR-m & 0.998 & 0.892 & 0.784 & 0.999  & 0.898 \\ \hline
			MLP-l & 0.998 & 0.892 & 0.784 & 0.999 & 0.896 \\ \hline
			MLP-l-m & 0.998 & 0.892 & 0.784 & 0.999 & 0.896 \\ \hline
			GBT & 0.998 & 0.833 & 0.667 & 1.000 & 0.856 \\ \hline
			GBT-m & 0.996 & 0.701 & 0.402 & 1.000  & 0.741 \\ \hline
			MLP-A & 0.996 & 0.652 & 0.304 & 1.000  & 0.700 \\ \hline
			MLP-A-m & 0.996 & 0.652 & 0.304 & 1.000 & 0.700 \\ \hline
			RF & 0.994 & 0.534 & 0.069 & 1.000 & 0.572 \\ \hline
			RF-m & 0.994 & 0.534 & 0.069 & 1.000  & 0.572 \\ \hline
			NB & 0.006 & 0.500 & 1.000 & 0.000 & 0.301 \\ \hline
		\end{tabular}
	\end{center}
\end{table}

\begin{table} [htb]
	\caption{Performance of Individual Models and Ensembles for the Turkish Transactions Database\cite{yap2020turca}.}
	\label{tab_result_ensemble_turca}
	\begin{center}
		\begin{tabular}{|l|l|l|l|l|l|}
			\hline
			Model/Ensemble & acc & bcr & sens & spec & Mean \\ \hline
			CC-MV 1572 & 0.873 & 0.853 & 0.833 & 0.873 & 0.858 \\ \hline
			CC-MV 1258 & 0.925 & 0.852 & 0.778 & 0.925 & 0.870 \\ \hline
			CC-MV 1559 & 0.925 & 0.852 & 0.778 & 0.925 & 0.870 \\ 
			\hline
			MLP-A & 0.902 & 0.840 & 0.778 & 0.902 & 0.855 \\ \hline
			MLP-A-m & 0.902 & 0.840 & 0.778 & 0.902 & 0.855 \\ \hline
			CC-OR 73 & 0.902 & 0.840 & 0.778 & 0.902 & 0.855 \\ \hline
			CC-OR 54 & 0.902 & 0.840 & 0.778 & 0.902 & 0.855 \\ \hline
			CC-OR 55 & 0.902 & 0.840 & 0.778 & 0.902 & 0.855 \\ \hline
			MLP-l & 0.683 & 0.786 & 0.889 & 0.683 & 0.760 \\ \hline
			MLP-l-m & 0.719 & 0.720 & 0.722 & 0.719 & 0.720 \\ \hline
			LR & 0.826 & 0.663 & 0.500 & 0.826 & 0.704 \\ \hline
			LR-m & 0.826 & 0.663 & 0.500 & 0.826 & 0.704 \\ \hline
			GBT & 0.864 & 0.654 & 0.444 & 0.864 & 0.707 \\ \hline
			KNN & 0.942 & 0.638 & 0.333 & 0.943 & 0.714 \\ \hline
			KNN-m & 0.942 & 0.638 & 0.333 & 0.942 & 0.714 \\ \hline
			GBT-m & 0.986 & 0.632 & 0.278 & 0.986 & 0.721 \\ \hline
			NB & 0.365 & 0.599 & 0.833 & 0.365 & 0.541 \\ \hline
			RF & 0.999 & 0.500 & 0.000 & 1.000 & 0.625 \\ \hline
			RF-m & 0.999 & 0.500 & 0.000 & 1.000 & 0.625 \\ \hline
		\end{tabular}
	\end{center}
\end{table}

In Table~\ref{tab_result_ensemble_europeia}, one may verify that the LR, MLP-l, KNN, and KNN-m models provided CCRs of up to 79\% for fraud (sens) and 89\% for both classes on average (bcr). The performance was inferior for GBT models, which reached CCRs of 66\% for fraud and 83\% for both classes on average. The other models exhibited bad performances, with CCRs for fraud smaller than 40\%. For the Turkish database (Table~\ref{tab_result_ensemble_turca}), in turn, NN models had the best performances, with CCRs between 72\% and 88\% for fraud and 82\% e 84\% for both classes on average. The remaining models exhibited bad performances, with CCRs between 0\% and 50\% for fraud. 

For both databases, the ensembles exhibited good performances. For the European database, there was an improvement of approximately 3pp (percentage points) in the fraud CCR and 2pp in the average CCR when all ensembles are compared to the best individual model (in this case, the KNN). For the Turkish database, the ensembles exhibited performances that were equal or exceeded in 5pp the fraud detection performance of the MLP-l models.

According to Table~\ref{tab_result_ensemble_europeia}, our proposed CC-OR strategy (Section~\ref{novo_metodoensemble_funcaoOR}) has a performance equivalent to that of the state-of-the-art CC-MV while incurring a lower computational cost. For the Turkish database (Table~\ref{tab_result_ensemble_turca}), the CC-OR had a CCR inferior in 1pp for both classes on average compared to the CC-MV. Nevertheless, in this case, the CC-OR outperformed the NN and all other individual models.

It is worth noticing that, for imbalanced data, the value of accuracy asymptotically converges to the value of specificity \cite{awoyemi2017ml} since the more frequent the majority class, the total number of correct classifications approaches that of the majority class.

\subsubsection{Mixed Learning using Individual Models and Ensembles} \label{results_mixed_learning}

In this section, we report the results of experiments in which we evaluated the performance of mixed learning techniques using, as supervised classifiers, both individual models and ensembles (Algorithms~\ref{alg5} and~\ref{alg6}). Tables~\ref{tab_result_ensemble_europeia_cluster} and~\ref{tab_result_ensemble_turca_cluster}, respectively, display the results obtained using the validation sets of the European and Turkish databases of Section~\ref{dtset}.

\begin{table} [htb]
	\caption{Performance after \textit{K-means} Clustering ($K=5$) for Individual Models and Ensembles for the European Transactions Database~\cite{mlgroup2017fraud}.}
	\label{tab_result_ensemble_europeia_cluster}
	\begin{center}
		\begin{tabular}{|l|l|l|l|l|l|}
			\hline
			Model/Ensemble & acc & bcr & sens & spec & Mean \\ \hline
			CC-MV 8 & 0.994 & 0.924 & 0.853 & 0.995 & 0.941 \\ \hline
			CC-MV 9 & 0.994 & 0.924 & 0.853 & 0.995 & 0.941 \\ \hline
			CC-MV 16 & 0.994 & 0.924 & 0.853 & 0.995 & 0.941 \\ \hline
			CC-OR 14 & 0.993 & 0.923 & 0.853 & 0.994 & 0.941 \\ \hline
			CC-OR 15 & 0.993 & 0.923 & 0.853 & 0.994 & 0.941 \\ \hline
			CC-OR 24 & 0.993 & 0.923 & 0.853 & 0.994 & 0.941 \\  \hline
			LR & 0.995 & 0.915 & 0.833 & 0.996 & 0.935 \\ \hline
			LR-m & 0.995 & 0.915 & 0.833 & 0.996 & 0.935 \\ \hline
			MLP-A & 0.971 & 0.912 & 0.853 & 0.972 & 0.927 \\ \hline
			MLP-A-m & 0.971 & 0.912 & 0.853 & 0.972 & 0.927 \\ \hline
			KNN & 0.995 & 0.910 & 0.824 & 0.996 & 0.931 \\ \hline
			KNN-m & 0.995 & 0.910 & 0.824 & 0.996 & 0.931 \\ \hline
			MLP-l & 0.968 & 0.901 & 0.833 & 0.969 & 0.918 \\ \hline
			MLP-l-m & 0.968 & 0.901 & 0.833 & 0.969 & 0.918 \\ \hline
			GBT-m & 0.941 & 0.898 & 0.853 & 0.942 & 0.911 \\ \hline
			GBT & 0.909 & 0.716 & 0.520 & 0.912 & 0.764 \\ \hline
			RF & 0.996 & 0.647 & 0.294 & 1.000 & 0.734 \\ \hline
			RF-m & 0.996 & 0.647 & 0.294 & 1.000 & 0.734 \\ \hline
			NB & 0.008 & 0.501 & 1.000 & 0.002 & 0.378 \\ \hline
		\end{tabular}
	\end{center}
\end{table}

\begin{table} [htb]
	\caption{Performance after \textit{K-means} Clustering ($K=5$) for Individual Models and Ensembles for the Turkish Transactions Database~\cite{yap2020turca}.}
	\label{tab_result_ensemble_turca_cluster}
	\begin{center}
		\begin{tabular}{|l|l|l|l|l|l|}
			\hline
			Model/Ensemble & acc & bcr & sens & spec & Mean \\ \hline
			RF & 0.999 & 0.999 & 1.000 & 0.999 & 0.999 \\ \hline
			RF-m & 0.999 & 0.999 & 1.000 & 0.999 & 0.999 \\ \hline
			CC-OR 51 & 0.999 & 0.999 & 1.000 & 0.999 & 0.999 \\ \hline
			CC-MV 1311 & 0.999 & 0.999 & 1.000 & 0.999 & 0.999 \\ \hline
			CC-MV 1366 & 0.999 & 0.999 & 1.000 & 0.999 & 0.999 \\ \hline
			CC-MV 1411 & 0.999 & 0.999 & 1.000 & 0.999 & 0.999 \\ \hline
			CC-OR 17 & 0.991 & 0.995 & 1.000 & 0.990 & 0.994 \\ \hline
			CC-OR 172 & 0.991 & 0.995 & 1.000 & 0.990 & 0.994 \\ \hline
			KNN-m & 0.990 & 0.995 & 1.000 & 0.990 & 0.994 \\ \hline
			CC-OR 13 & 0.990 & 0.995 & 1.000 & 0.990 & 0.994 \\ \hline
			GBT-m & 0.989 & 0.994 & 1.000 & 0.988 & 0.993 \\ \hline
			KNN & 0.991 & 0.993 & 0.995 & 0.991 & 0.992 \\ \hline
			GBT & 0.942 & 0.951 & 0.961 & 0.942 & 0.949 \\ \hline
			MLP-lbfgs & 0.878 & 0.921 & 0.966 & 0.876 & 0.910 \\ \hline
			MLP-l-m & 0.878 & 0.921 & 0.966 & 0.876 & 0.910 \\ \hline
			MLP-A & 0.932 & 0.894 & 0.854 & 0.934 & 0.904 \\ \hline
			MLP-A-m & 0.932 & 0.894 & 0.854 & 0.934 & 0.904 \\ \hline
			LR & 0.854 & 0.825 & 0.794 & 0.856 & 0.832 \\ \hline
			LR-m & 0.854 & 0.825 & 0.794 & 0.856 & 0.832 \\ \hline
			NB & 0.739 & 0.822 & 0.909 & 0.735 & 0.801 \\ \hline
		\end{tabular}
	\end{center}
\end{table}

As one may observe, the mixed learning techniques led to improved performance. For the European database, comparing the results in Tables~\ref{tab_result_ensemble_europeia} and~\ref{tab_result_ensemble_europeia_cluster}, one may verify increases in CCRs of up to 6pp for fraud events and 2pp on average. For the Turkish database, in turn, comparing Tables~\ref{tab_result_ensemble_turca} and~\ref{tab_result_ensemble_turca_cluster}, one may notice a more significant performance gain: using mixed learning, four models exhibit CCRs of 100\% for fraud and 99\% for genuine transactions, and all remaining models have CCRs superior to 79\% for fraud. Without clustering preprocessing, the best classifier has CCRs of 77\% for fraud and 90\% for genuine transactions.

In the mixed learning setups, the ensembles' performances also exceeded that of the individual models, as observed in Section~\ref{results_ensembles}. For the European database, the best performing ensembles were better than the best individual model (KNN) in 3pp in fraud CCR while keeping a CCR of 99\% for genuine transactions. For the Turkish database, there was an increase of 6pp in the detection rate of fraudulent transactions but a decline of 3pp in the detection rate of genuine ones, which nonetheless may be advantageous from a financial point of view. Finally, we point out that the proposed CC-OR strategy (Section~\ref{novo_metodoensemble_funcaoOR}) also exhibited a performance equivalent to that of the state-of-the-art CC-MV technique when deployed in mixed learning setups while incurring a lower computational cost.

In Table~\ref{tab_lista_ensembles} we describe the compositions of the best performing ensembles listed in Tables~\ref{tab_result_ensemble_europeia}, \ref{tab_result_ensemble_turca}, \ref{tab_result_ensemble_europeia_cluster}, and~\ref{tab_result_ensemble_turca_cluster}.

\begin{table} [htb]
	\caption{Composition of Ensembles (\textit{Ensemble})}
	\label{tab_lista_ensembles}
	\begin{center}
		\begin{tabular}{|l|l|}
	\hline
	\textit{Ensemble} & Composition \\ \hline
CC-MV 8 & NB\hspace{0.05cm} KNN\hspace{0.05cm} KNN-m\hspace{0.05cm} LR\hspace{0.05cm} MLP-l \\ \hline
CC-MV 9 & NB\hspace{0.05cm} KNN\hspace{0.05cm} KNN-m\hspace{0.05cm} LR\hspace{0.05cm} MLP-l-m \\ \hline
CC-MV 16 & NB\hspace{0.05cm} KNN\hspace{0.05cm} KNN-m\hspace{0.05cm} LR-m\hspace{0.05cm} MLP-l \\ \hline
CC-MV 1258 & RF\hspace{0.05cm} GBT\hspace{0.05cm} MLP-A\hspace{0.05cm} MLP-A-m\hspace{0.05cm} MLP-l \\ \hline
CC-MV 1311 & NB\hspace{0.1cm} RF\hspace{0.1cm} RF-m \\ \hline
CC-MV 1366 & KNN\hspace{0.1cm} RF\hspace{0.1cm} RF-m \\ \hline
CC-MV 1411 & KNN-m\hspace{0.1cm} RF\hspace{0.1cm} RF-m \\ \hline
CC-MV 1559 & RF-m\hspace{0.02cm} GBT\hspace{0.02cm} MLP-A\hspace{0.02cm} MLP-A-m\hspace{0.02cm} MLP-l \\ \hline
CC-MV 1572 & GBT\hspace{0.02cm} MLP-A\hspace{0.02cm} MLP-A-m\hspace{0.02cm} MLP-l\hspace{0.02cm} MLP-l-m \\ \hline
CC-OR 13 & KNN\hspace{0.1cm} KNN-m \\ \hline
CC-OR 14 & KNN\hspace{0.1cm} LR \\ \hline
CC-OR 15 & KNN\hspace{0.1cm} LR-m \\ \hline
CC-OR 17 & KNN\hspace{0.1cm} RF-m \\ \hline
CC-OR 24 & KNN-m\hspace{0.1cm} LR \\ \hline
CC-OR 51 & RF\hspace{0.1cm} RF-m \\ \hline
CC-OR 172 & KNN\hspace{0.1cm} RF\hspace{0.1cm} RF-m \\ \hline
		\end{tabular}
\end{center}
\end{table}

\section{Conclusions}
\label{conclusao}

In this work, we proposed the CC-OR, CCK-OR, and CCK-VM algorithms as new approaches to tackle the problem of fraud detection fraud in credit card transactions. We performed simulations using databases with real-world credit card transactions. From their results, we observed that ensembles perform equivalently or better than the best-performing individual models. We also observed that the proposed CC-OR aggregation strategy performed equivalently to the state-of-the-art CC-MV method while incurring a lower computational complexity. Additionally, we verified that the mixed learning technique that performs clustering on data before supervised classification improved performance when deployed with both individual models and ensembles. We believe, therefore, that the proposed algorithms are viable alternatives to present fraud detection schemes.

\bibliography{IEEEabrv.bib,Draft_English_v6.bib}{}
\bibliographystyle{IEEEtran}

\begin{IEEEbiography}{Daniel H. M. de Souza} 
received both the B.S. degree in Management Engineering and the M.S. degree in Information Engineering from Universidade Federal do ABC (UFABC), Santo Andr\'{e}, Brazil, in 2008. Currently, he is pursuing a Ph.D. degree in Information Engineering at UFABC. He works as Data Science Coordinator at Grupo Botic\'{a}rio. He is also a lecturer in graduate programs in Artificial Intelligence and Data Analysis at Instituto Mau\'{a} de Tecnologia (IMT) and at Faculdade de Tecnologia Termomec\^{a}nica (FTT). His research interests lie in the areas of Applied Statistics, Artificial Intelligence, Machine Learning, Data Engineering, and Big Data.
\end{IEEEbiography}	

\begin{IEEEbiography}{Claudio J. Bordin Jr.} (M'05)
received the B.S., M.S., and Ph.D. degrees, all in Electrical Engineering, from Escola Politécnica, Universidade
de São Paulo, Brazil, in 2000, 2002, and 2006, respectively. He is currently an Associate Professor at Universidade Federal do ABC, Santo André, Brazil. His research interests lie in the areas of Signal Processing, Applied Statistics, and Machine Learning.	 
\end{IEEEbiography}

\end{document}